\documentclass[journal]{IEEEtran}
\usepackage{amsmath,amsfonts}
\usepackage{array}
\usepackage[caption=false,font=normalsize,labelfont=sf,textfont=sf]{subfig}
\usepackage{textcomp}
\usepackage{stfloats}
\usepackage{url}
\usepackage{verbatim}
\usepackage{graphicx}
\usepackage{cite}
\usepackage{amssymb}
\usepackage{bm}
\usepackage{xcolor}

\usepackage{multirow}
\usepackage{amstext}
\usepackage{algorithm}
\usepackage{algpseudocode}
\usepackage{bbding}
\usepackage[inline]{enumitem}
\usepackage{makecell}
\usepackage{booktabs}
\definecolor{fqr}{RGB}{35,146,144}
\definecolor{fqf}{RGB}{242,70,57}

\def\etal{\emph{et al}.}

\def\ie{\emph{i.e}.}
\newcommand{\fq}[1]{{{#1}}}
\newcommand{\yl}[1]{{{#1}}}
\newcommand{\yyw}[1]{{#1}}
\newcommand{\yywy}[1]{{#1}}

\usepackage{hyperref}

\begin{document}

\title{FOF-X: Towards Real-time Detailed Human Reconstruction from a Single Image}

\author{Qiao Feng$^\dagger$, ~\IEEEmembership{Student Member,~IEEE,} Yuanwang Yang$^\dagger$, ~\IEEEmembership{Student Member,~IEEE,} Yebin Liu, ~\IEEEmembership{Member,~IEEE,} Yu-Kun Lai, ~\IEEEmembership{Senior Member,~IEEE,} Jingyu Yang, ~\IEEEmembership{Senior Member,~IEEE,} and Kun Li$^*$, ~\IEEEmembership{Senior Member,~IEEE}
\thanks{$^\dagger$ Equal contribution.}
\thanks{$^*$ Corresponding author: Kun Li (E-mail: lik@tju.edu.cn).}
\thanks{This work was partially supported by the National Key R\&D Program of China (2023YFC3082100) and the National Natural Science Foundation of China (62171317, 62125107, and 62231018).}
\thanks{Qiao Feng, Yuanwang Yang and Kun Li are with the College of Intelligence and Computing, Tianjin University, Tianjin 300350, China. E-mail: \{fengqiao,yyw,lik\}@tju.edu.cn}
\thanks{Yebin Liu is with the Department of Automation, Tsinghua University, Beijing 100084, China. E-mail: liuyebin@mail.tsinghua.edu.cn}
\thanks{Yu-Kun Lai is with the School of Computer Science and Informatics, Cardiff University, Cardiff CF24 4AG, United Kingdom. E-mail: Yukun.Lai@cs.cardiff.ac.uk}
\thanks{Jingyu Yang is with the School of Electrical and Information Engineering, Tianjin University, Tianjin 300072, China. E-mail: yjy@tju.edu.cn}
}



\maketitle

\begin{abstract} 
We introduce FOF-X for real-time reconstruction of detailed human geometry from a single image. 
Balancing real-time speed against high-quality results is a persistent challenge, mainly due to the high computational demands of existing 3D representations. To address this, we propose Fourier Occupancy Field (FOF), an efficient 3D representation by learning the Fourier series. 
The core of FOF is to factorize a 3D occupancy field into a 2D vector field, retaining topology and spatial relationships within the 3D domain while facilitating compatibility with 2D convolutional neural networks. 
Such a representation bridges the gap between 3D and 2D domains, enabling the integration of human parametric models as priors and enhancing the reconstruction robustness.
Based on FOF,  we design a new reconstruction framework, FOF-X, to avoid the performance degradation caused by texture and lighting. This enables our real-time reconstruction system to better handle the domain gap between training images and real images.
Additionally, in FOF-X, we enhance the inter-conversion algorithms between FOF and mesh representations with a Laplacian constraint and an automaton-based discontinuity matcher, improving both quality and robustness. 
We validate the strengths of our approach on different datasets and real-captured data, where FOF-X achieves new state-of-the-art results. 
The code has already been released for research purposes at \href{https://cic.tju.edu.cn/faculty/likun/projects/FOFX/index.html}{\textcolor{magenta}{https://cic.tju.edu.cn/faculty/likun/projects/FOFX/index.html}}.
\end{abstract}


\section{Introduction}
\IEEEPARstart{R}{econstructing} a 3D human from a single image has emerged as a popular task in computer vision and graphics, which can be widely used in various downstream applications, such as mixed reality and virtual try-on. However, real-time, high-fidelity monocular 3D human reconstruction remains challenging. The core of this challenge lies in the 3D representation, as it significantly influences the design and performance of reconstruction and generation approaches. 
Despite \yyw{some} promising results,
existing methods~\cite{saito2019pifu, sifu,xiu2022icon, econ, liu2025learning, mao2025deep} typically suffer from high computational consumption and lack of robustness.
Consequently, a good 3D geometry representation is essential for 3DTV, Holographic Telepresence systems, and other real-time applications, which must fulfill the requirements of accuracy, efficiency, and compatibility \cite{alatan20073DTV,yang2023high}.

\begin{figure}[!t]
\centering
\includegraphics[width=0.98\linewidth]{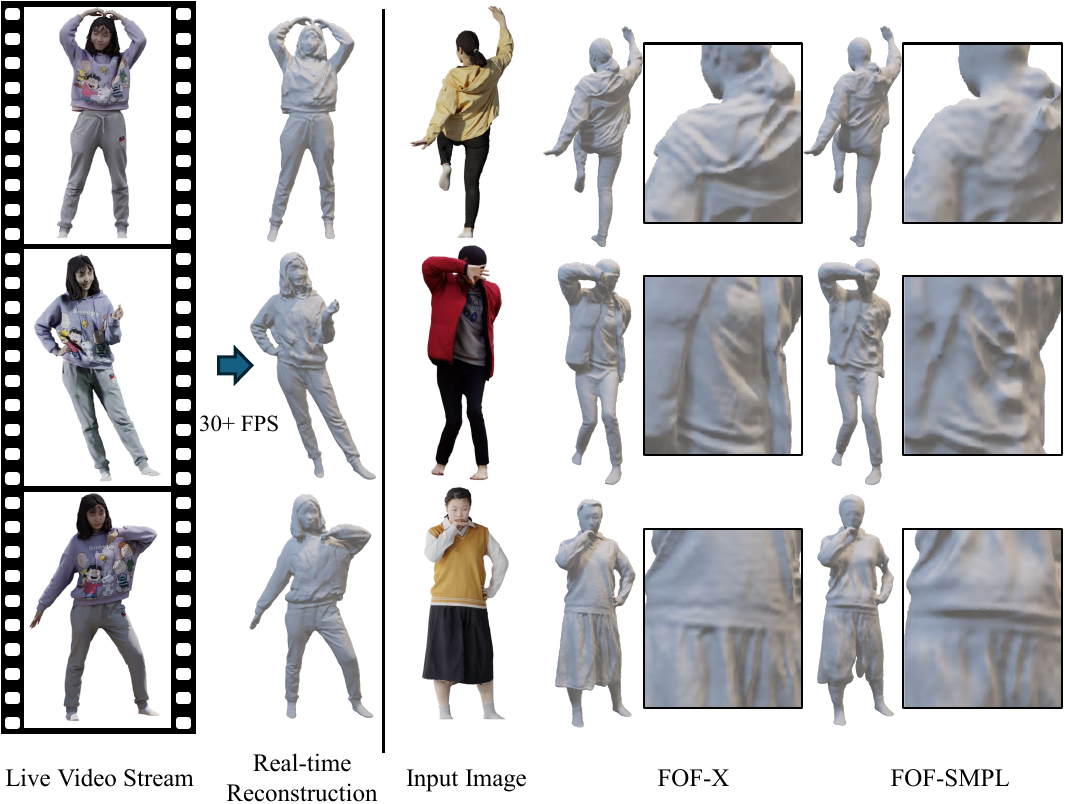}
\caption{Left: Our FOF-X can reconstruct 3D human shapes from a live video stream with a real-time speed of over 30 FPS. Right: Compared to the original FOF-SMPL, our FOF-X demonstrates better robustness to variations in texture and lighting. Under challenging lighting conditions, such as strong illumination or shadows, FOF-X produces more detailed and accurate reconstructions (first two rows). FOF-X effectively avoids the incorrect reconstruction caused by textures, such as the stripe pattern on the edge of the shirt (third row). Note that FOF-SMPL is not real-time.}
\label{teaser}
\end{figure}
Classic representations, such as voxel grids \cite{zheng2019deephuman} or meshes \cite{zhu2021HMD,alldieck2019tex2shape,li2021TIP}, have been explored for monocular human reconstruction. However, voxel grids require a space complexity of $O(n^3)$ (where $n$ is the resolution of grids in each dimension), and meshes struggle with topology changes or large deformations.
Although some recent methods \cite{econ, tech} propose optimization-based pipelines combining normal maps, voxel grids, and meshes, they all suffer from efficiency issues.

Implicit neural representations have emerged and been widely used in monocular human reconstruction \cite{saito2019pifu,saito2020pifuhd,zheng2021pamir,alldieck2022phorhum,xiu2022icon}. These methods treat 3D space as a continuous field, such as an occupancy or signed distance field, represented as $F(x,y,z):\mathbb{R}^3 \to \mathbb{R}$. Instead of voxel grids, a neural network models the field, allowing results at any resolution. However, \yyw{these methods} infer values across numerous spatial grid points, which slows down inference and makes high-frame-rate or real-time reconstruction a significant challenge.
Based on PIFu \cite{saito2019pifu}, Monoport \cite{li2020monoport} proposes an efficient sampling scheme to speed up inference, achieving 15 FPS only for mesh-free rendering. However, when generating a
complete mesh, their method operates at less than 10 FPS, and the results are often worse than PIFu. These limitations hinder \yyw{Monoport's} applicability with downstream applications. 

\begin{table}[!t]
  \scriptsize
  \caption{Comparison with existing 3D representations}
  \label{table1}
  \centering
  \resizebox{\linewidth}{!}{
  \begin{tabular}{lcccc}
    \toprule
    Representation    & \makecell[c]{Aligned\\with Images} &  
    \makecell[c]{High-Quality}&
    \makecell[c]{Computational\\Efficiency} &
    \makecell[c]{Flexible\\Topology}\\
    \midrule
    \makecell[l]{Parametric Model\\ \quad
    \cite{SMPL:2015,SMPL-X:2019,STAR:2020}}  & \color{fqf}\XSolidBrush & \color{fqf}\XSolidBrush &
    \color{fqr}\Checkmark &  \color{fqf}\XSolidBrush \\
    Voxel Grid\cite{zheng2019deephuman}  & \color{fqr}\Checkmark & \color{fqf}\XSolidBrush &  \color{fqf}\XSolidBrush  &  \color{fqr}\Checkmark\\
    Mesh\cite{alldieck2019tex2shape, li2021TIP, zhu2021HMD}  & \color{fqf}\XSolidBrush & \color{fqf}\XSolidBrush & \color{fqr}\Checkmark  &\color{fqf}\XSolidBrush \\
    \makecell[l]{Implicit Function\cite{saito2019pifu}, \\ \quad \cite{saito2020pifuhd,zheng2021pamir,alldieck2022phorhum,xiu2022icon,sith}}    & \color{fqr}\Checkmark & \color{fqr}\Checkmark & \color{fqf}\XSolidBrush &  \color{fqr}\Checkmark\\
    Our FOF(-X) & \color{fqr}\Checkmark & \color{fqr}\Checkmark & \color{fqr}\Checkmark &  \color{fqr}\Checkmark\\
    \bottomrule
  \end{tabular}}
\end{table}

\begin{figure}[!t]
\centering
\includegraphics[width=0.95\linewidth]{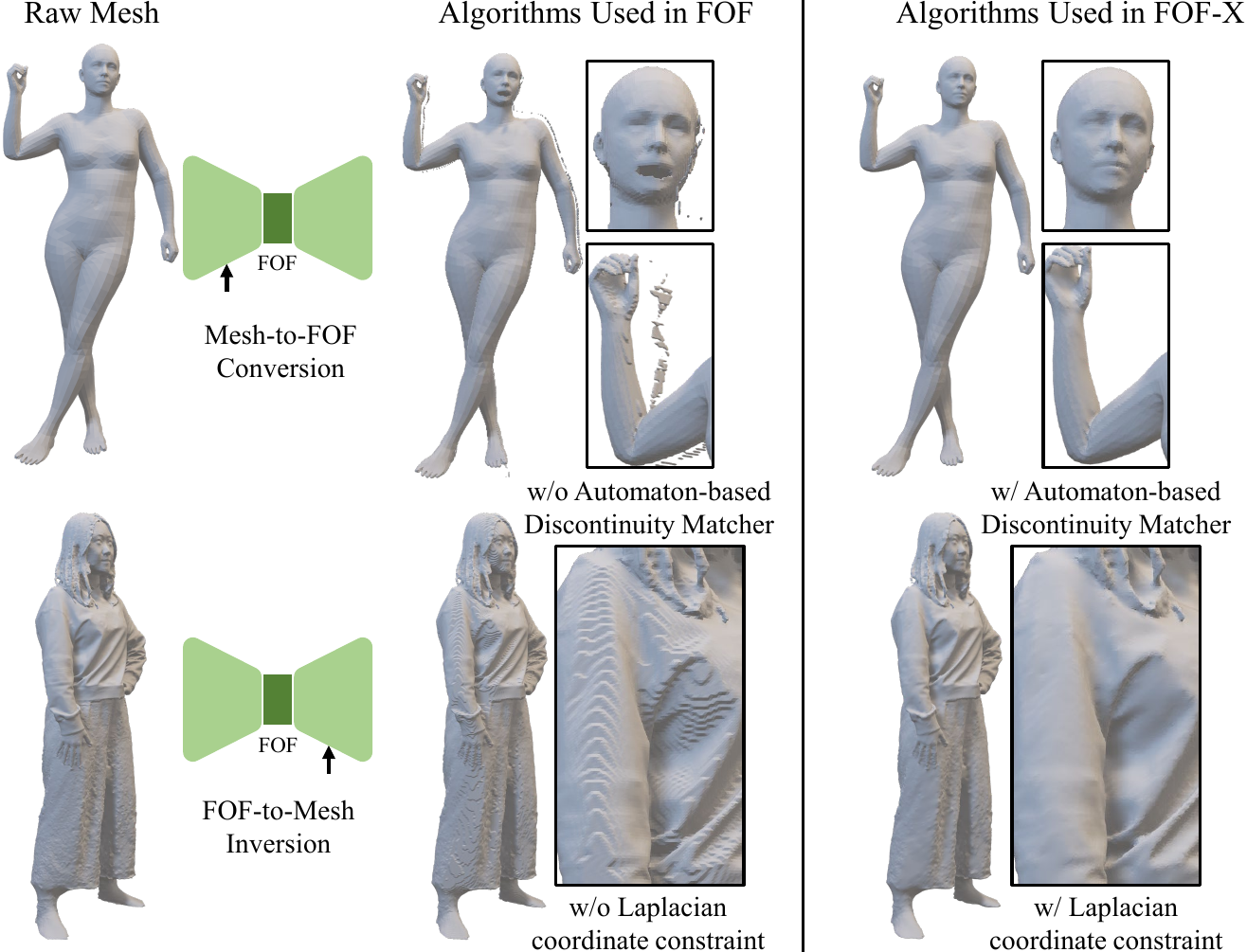}
\caption{FOF and meshes can be inter-converted flexibly. The newly designed inter-conversion algorithms in FOF-X exhibit better robustness and quality. Our automaton-based discontinuity matcher eliminates floating artifacts during the conversion process (first row). With the Laplacian coordinate constraint, we resolve stair-step artifacts on the recovered meshes (second row).}
\vspace{-0.5cm}
\label{imp}
\end{figure}

To address these challenges, we propose \emph{Fourier Occupancy Field (FOF)}, a novel representation for monocular real-time and detailed 3D human reconstruction. FOF is an expressive, efficient, and flexible 3D geometric representation manifested as a 2D map aligned with the input image. The key idea of FOF is to represent a 3D object with a 2D field by decomposing the occupancy field along the $z$ axis into a Fourier series, retaining only the first few terms. This approach is not only memory-efficient but also adept at preserving essential geometric information. 
By implementing grid sampling on the $x$ and $y$ axes, FOF can be explicitly stored as a multi-channel image.
This feature allows 3D geometric priors, such as SMPL models, to be fed directly into the CNN. With our inter-conversion algorithms between FOF and mesh representations, 3D and 2D information can be processed within a unified framework.
Unlike depth maps, FOF encapsulates the complete geometry of an object, not just its visible parts, significantly enhancing the fidelity of the reconstructed human shapes. 
Detailed comparisons with existing representations are presented in Table \ref{table1}. 

When applying FOF directly to a real-time system, we observed certain limitations in robustness. The reconstruction is degraded by texture and lighting effects. Additionally, efficiency and robustness issues in the original mesh-to-FOF conversion algorithm used in the conference version prevent the use of the SMPL prior in the real-time system, resulting in unsatisfactory performance on challenging poses. Furthermore, the original FOF-to-mesh algorithm comes with stair-step artifacts, further compromising the reconstruction quality.
To address these challenges, we propose FOF-X, a novel reconstruction framework. With a single RGB image as input, FOF-X first translates it to dual-sided normal maps, which are unaffected by texture and lighting, as an internal representation of the human body. \yyw{This strategy} allows the network to focus on geometric details, significantly improving the robustness and performance of the reconstruction.
\yyw{To ensure the robustness and fidelity of FOF computations, we design an automaton-based discontinuity matcher to filter out \yyw{invalid} fragments.} This matcher can be parallelized on a GPU, enabling real-time conversion of parametric models. Additionally, we incorporate a Laplacian coordinate constraint in the FOF-to-mesh conversion process to eliminate artifacts caused by view bias, resulting in more accurate mesh reconstructions and improving the fidelity of the final output.
As shown in Fig. \ref{imp}, the newly designed mesh-to-FOF and FOF-to-mesh algorithms in FOF-X effectively address the robustness issues.
Based on these updates, our final method, FOF-X, achieves new state-of-the-art results in both speed and accuracy, contributing a novel solution for real-time detailed human reconstruction from a single image. Experiments also show that FOF-X has better generalization for real-captured data.



The main contributions of our work are as follows:
\begin{itemize}
\item We propose \emph{Fourier Occupancy Field (FOF)}, a novel representation for 3D humans, which can represent a high-quality geometry with a 2D map aligned with the image, bridging the gap between 2D images and 3D geometries. 

\item We introduce FOF-X, a new reconstruction framework designed for detailed 3D human reconstruction from a single RGB image. FOF-X utilizes dual-sided normal maps as an internal representation, avoiding the effects of texture and lighting to focus on geometric accuracy.

\item In FOF-X, we design parallelized inter-conversion algorithms between FOF and meshes with a Laplacian coordinate constraint and an automaton-based discontinuity matcher, further enhancing the \yyw{robustness and reconstruction quality}.

\item Compared with state-of-the-art methods based on other representations, our approach can produce high-quality results in real-time. Our system is the first 30+FPS pipeline, achieving a twofold improvement in speed over Monoport \cite{li2020monoport}, coupled with superior quality.
\end{itemize}

An early version of our method, including only FOF, was published as a conference paper \cite{fof}. \yyw{Based on our FOF representation, several works for generation\cite{joint2human} and reconstruction\cite{li2024diffusion, yang2024r2human,yang2024real} have been proposed. 
In this paper, we substantially extend the original version by developing \textbf{FOF-X}. Firstly, we design a new reconstruction framework that greatly mitigates the performance degradation caused by texture and lighting effects.
Secondly, we propose a robust mesh-to-FOF conversion algorithm with an automaton-based discontinuity matcher, enabling real-time execution and significantly improving the system's robustness when facing challenging poses. 
Thirdly, we \yyw{propose a} FOF-to-mesh algorithm with a Laplacian coordinate constraint for greater robustness and fidelity. Such a strategy effectively addresses artifacts caused by view-direction bias without any loss of geometric details. 
Fourth, we employ cosine series as the subspace approximation basis, retaining the Fourier-like approximation capacity while eliminating Gibbs phenomena through even-periodic boundary extension.  Additionally, we evaluate FOF-X through comprehensive experiments including: testing on more datasets, comparisons with various state-of-the-art methods, and more detailed ablation studies, all confirming the method's effectiveness, efficiency and robustness.}

\yyw{FOF-X offers three key advantages:
(1) FOF-X is sampling-scalable along all three spatial axes, allowing for resolution adaptation at inference without retraining, which supports diverse deployment scenarios with varying speed, quality, and memory constraints;
(2) FOF-X represents 3D objects as multi-channel images, requiring only a simple tensor multiplication to reconstruct the 3D occupancy field;
(3) FOF-X can be seamlessly inter-converted to and from the mesh representation, making it compatible with traditional \yyw{graphics pipelines}.
Additionally, as a compact and computationally efficient 3D representation, FOF-X is suitable for real-time applications \yyw{like holographic transportation.}}

\section{Related Work}
Monocular 3D human reconstruction methods can be grouped into surface-based, volume-based, and ``sandwich-like'' approaches. In addition, recent works explore generative models for completion from a single image.

\subsection{Surface-based Reconstruction}
Surface-based approaches focus on inferring the interface between geometry and empty space, and can be broadly classified into three categories: parametric models, UV-map based methods, and graph-based methods.
Parametric models such as SMPL \cite{SMPL:2015}, SMPL-X \cite{SMPL-X:2019}, and STAR \cite{STAR:2020} represent naked human bodies via statistical shape spaces and are widely used to regress meshes from RGB or RGBD images \cite{pymaf2021,hu20213dbodynet}. Some methods~\cite{tang2023mlp, yu2023mobirfpose} focus solely on 3D human pose estimation, and their outputs can also be converted into parametric models. However, they cannot capture clothing or hair.
UV-map-based methods deform parametric models by estimating per-vertex displacements. Zhu \etal~\cite{zhu2019HMD1,zhu2021HMD} employ a four-stage process constrained by joints, silhouettes, and shading,
but their results are sometimes inconsistent with the images.
Graph-based representations are naturally compatible with triangle meshes. Li \etal~\cite{li2021TIP} employ graph neural networks for topology-consistent reconstructions, but results are smooth and lack details. Habermann \etal~\cite{Deecap2} capture dense performance from monocular video but require pre-scanned templates. Overall, surface-based methods struggle with topology variation and fine detail recovery.

\subsection{Volume-based Reconstruction}
Volume-based methods predict occupancy or signed distance values, naturally handling arbitrary topology.
Early voxel-based CNNs \cite{zheng2019deephuman} are memory-heavy and low-resolution. Implicit representations overcome this, as in PIFu \cite{saito2019pifu} and PIFuHD \cite{saito2020pifuhd}, which regress pixel-aligned functions and normals for high-detail geometry. PAMIR \cite{zheng2021pamir} and ICON \cite{xiu2022icon} leverage SMPL priors for robustness, though outputs often misalign with image evidence and appear noisy.
Recent works enhance implicit fields with transformers \cite{gta,sifu}, but inference is computationally expensive. Li \etal~\cite{li2020monoport} improve sampling efficiency, yet real-time performance remains elusive.
Recent NeRF \cite{2020NeRF} has also been adapted for humans \cite{sherf,elicit}, enabling supervision via volume rendering. While SHERF \cite{sherf} generalizes across humans and ELICIT \cite{elicit} leverages CLIP, NeRF-based approaches mainly excel at view synthesis rather than accurate geometry. Meanwhile, GS-SFS~\cite{jiang2024gs} reconstructs humans via Gaussian Splatting but relies on dense view inputs and struggles with recovering fine geometric details.

\subsection{``Sandwich-like'' Reconstruction}
In recent works \cite{modulingH,gin,econ}, 3D objects are modeled using \yl{front and back} depth maps, resembling a sandwich structure that encloses the inner space. However, depth maps only represent part of the geometric surface and cannot fully reconstruct the object. Therefore, these methods require volume-based post-processing, such as IF-Nets \cite{if-nets} and Poisson surface reconstruction\cite{2006Poisson}, to complete the result further. Any-Shot GIN \cite{gin} extends the sandwich-like scheme to novel classes of objects by predicting the depth maps of two sides from a single image and performing shape completion with IF-Nets. ECON \cite{econ} follows a similar approach and extends it to 3D humans. It estimates the normal maps of two sides and converts them into depth maps using an optimization method. The parametric human model prior is also employed in the shape completion stage to ``inpaint'' the missing geometry. However, this optimization process is time-consuming, and its instability may lead to incorrect geometric estimates.

\subsection{Leveraging Generative Models}
Generative models address missing geometry by synthesizing plausible invisible regions.
SiTH \cite{sith} generates back views with diffusion to estimate dual-sided normals. TeCH \cite{tech} integrates DMTet with SDS \cite{poole2022dreamfusion} for SMPL-X mesh recovery, yielding visually appealing but noisy surfaces. Human-LRM \cite{humanlrm2023} combines NeRF-based LRM with diffusion-based refinement, though geometric gains are limited.
PSHuman \cite{li2025pshuman} produces multi-view images via diffusion and refines SMPL-X meshes with differentiable rasterization, but its heavy pipeline hampers efficiency. Text2Avatar \cite{kwon2025text2avatar} generates avatars from text instructions but still requires dense view inputs for high-quality results.
In general, generative methods are time-consuming and often fail to preserve input fidelity.

To solve the problems with existing representations, we propose the Fourier Occupancy Field (FOF), a representation that is both image-aligned and efficient. Building upon FOF, we design FOF-X and parallel inter-conversion algorithms, achieving robust high-quality monocular human reconstruction at 30 FPS.

\section{Method}

\begin{figure*}[!t]
\centering
\includegraphics[width=\linewidth]{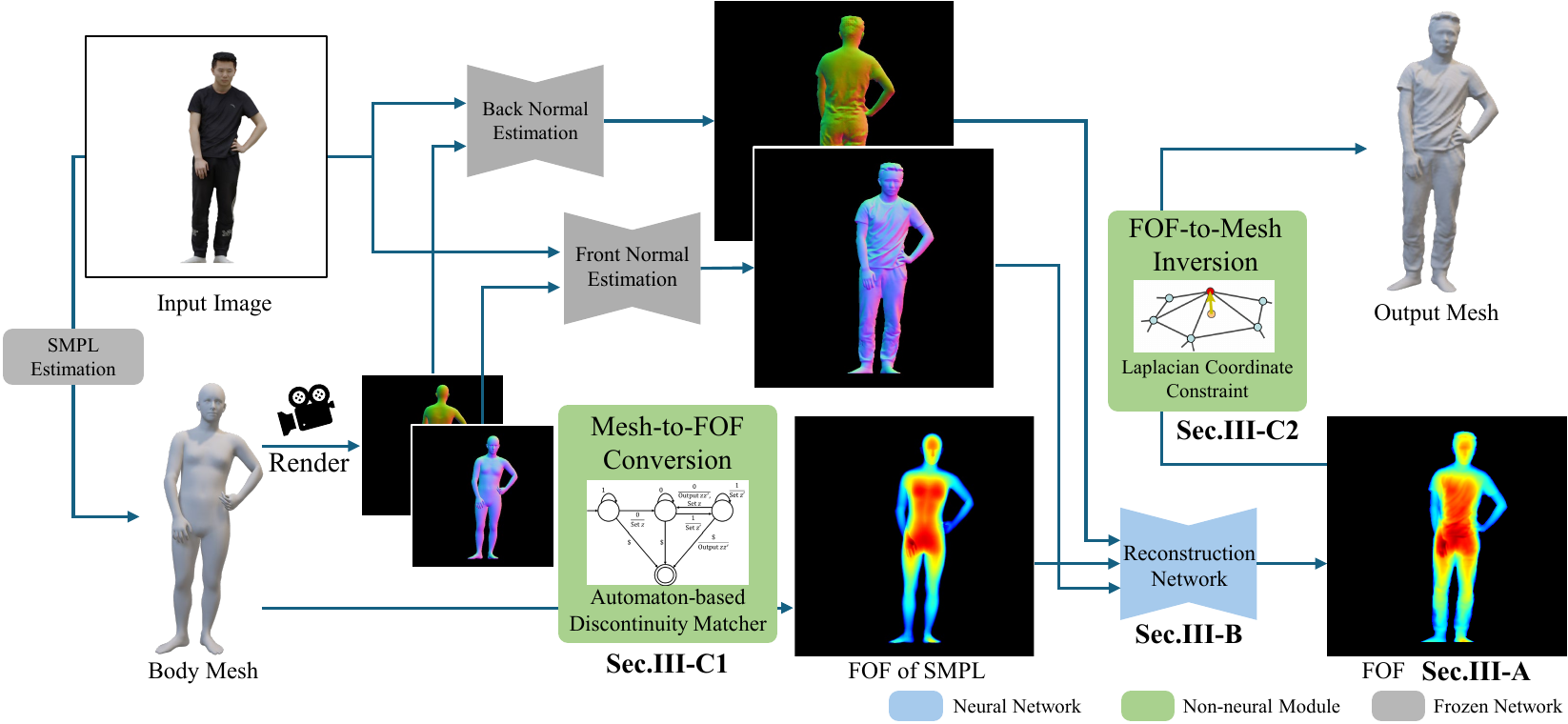}
\caption{The overall pipeline of FOF-X for monocular real-time human reconstruction. FOF-X takes an RGB image as input and exploits a SMPL body mesh as a prior with the proposed mesh-to-FOF conversion algorithm (Sec. \ref{convertion1}), which includes an automaton-based \yl{discontinuity} matcher to ensure robustness. 
Based on the rendered SMPL normal maps and input RGB image, the dual-sided normal maps are predicted as the internal representation and decoded to FOF with the SMPL prior through an image-to-image network (Sec. \ref{learn}). The FOF representation (Sec. \ref{formulation}) is then converted to \yl{a} mesh with the FOF-to-mesh inversion module (Sec. \ref{convertion2}), incorporating a Laplacian coordinate constraint to enhance the quality of the output mesh. }
\label{Networks}
\end{figure*}

\fq{Fig. \ref{Networks} illustrates the overall pipeline of our method. Our work aims to reconstruct a high-fidelity 3D human model from a single RGB image in \yl{real time}. In this section, we elaborate on the technical details of our approach. We first introduce our vanilla \emph{FOF (Fourier Occupancy Field)}, an efficient and flexible representation for 3D geometry, in Sec. \ref{formulation}. Then, we extend \emph{FOF} to \emph{FOF-X} with 
a new reconstruction framework with dual-sided normal maps as an internal representation  (Sec. \ref{learn}), a parallelized mesh-to-FOF conversion algorithm with an automaton-based discontinuity matcher (Sec. \ref{convertion1}), and a more robust FOF-to-mesh extraction algorithm with a Laplacian coordinate constraint (Sec. \ref{convertion2}). These updates further improve the quality of the reconstructed meshes while keeping the overall pipeline running in \yl{real time}. }

\subsection{Formulation of Fourier Occupancy Field}
\label{formulation}

\subsubsection{Vanilla FOF}
Given a normalized cube $[-1,1]^3$, the occupancy field $F: [-1,1]^3 \mapsto {0,0.5,1}$ is defined as
\begin{equation}
\label{equ:occu_fun}
F(x,y,z)=\left\{
  \begin{array}{ll}
    1, & \text{$(x,y,z)$ is inside the object},\\
    0.5, & \text{$(x,y,z) \in S$},\\
    0, & \text{$(x,y,z)$ is outside the object}.  
  \end{array}
\right.
\end{equation} 
Such a 3D representation is highly redundant, because only a small subset, \ie, the iso-surface of $F(x,y,z)$ with value of 0.5, is sufficient to represent the surface of the object. 
Direct inference of the 3D occupancy field $F$ from a single image not only confronts with the curse of dimensionality, but also requires more computation and memory in handling high dimensional feature maps. Note that the occupancy field defined in the 3D cube can be regarded as the collection of a large number of 1D signals defined on the lines along the view direction.
Without loss of generality, we assume that the view direction is the same as the $z$-axis of the 3D cube. The occupancy function, as a 1D signal $f(z): [-1, 1] \mapsto \left\{0,0.5,1\right\}$, along the line passing through a particular point $(x^*, y^*)\in [-1, 1]^2$ on the $xy$-plane can be written as $f(z) = F(x^*,y^*,z)$. We explore Fourier representation for such a family of 1D occupancy signals and propose a more compact 2D vector field orthogonal to the view direction, namely \emph{Fourier Occupancy Field (FOF)}, for more efficient representation of the 3D occupancy field, which is detailed in the following subsections.

{\bf{Fourier Series on a Single Occupancy Line:}} 1D occupancy signals $\left\{f(z)\right\}$ are essentially on-off signals switching at the boundaries of human bodies, which lie in a low-dimensional manifold in the ambient signal space. 
The $f_p(z)$ can be expanded as a convergent Fourier series:
\begin{equation}
  \label{FS}
  f_{p}(z)= \frac{a_0}{2} + \sum_{n=1}^{\infty}\left(a_n\cos (n\pi z)+b_n\sin (n\pi z)\right),
\end{equation} 
where $\{a_n\}, \{b_n\} \in\mathbb{R}$ are coefficients of basis functions  $\left\{ \cos(nx)\right\}$ and $\left\{ \sin(nx)\right\}$, respectively. 
Note that by defining the occupancy function as Eq. (\ref{equ:occu_fun}), Eq. (\ref{FS}) holds for discontinuity at $z^*$ on the surface $S$  because $f(z^*)=(f(z^*-)+f(z^*+))/2=0.5$, where $f(z^*-)$ and $f(z^*+)$ are the left hand limit and right hand limit of $f(z)$ at $z^*$. \yywy{For compact representation, we approximate 1D occupancy signals $\hat{f}(z)$ by a subspace spanned by the first $2N+1$ basis functions:}
\begin{equation}
  \hat{f}(z) = \bm{b}^\top(z) \bm{c},
\end{equation}
where $ \bm{b}(z)=[1/2,\cos(z),\sin(z),\ldots, \cos(Nz),\sin(Nz)]^\top$ is the vector of the first $2N+1$ basis functions spanning the approximation subspace, and $ \bm{c} = [a_0, a_1, b_1, \ldots, a_N, b_N]^\top$ is the coefficient vector which provides a more compact representation of the 1D occupancy function $f(z)$. 

{\bf{Fourier Occupancy Field:}} \yl{Such} a Fourier subspace approximation is applied to all 1D occupancy signals over the $xy$-plane:  
\begin{equation}
  \label{FOF}
  \hat{F}(x,y,z) = \bm{b}^\top(z) \bm{C}(x, y),
\end{equation}
where $\bm{C}(x,y)$ is the ($2N+1$)\yl{-dimensional} Fourier coefficient \yl{vector} for the 1D occupancy \yl{signal} at $(x, y)$. \yywy{In this way, we obtain the \emph{Fourier Occupancy Field} $\bm{C}:[-1,1]^2 \mapsto \mathbb{R}^{2N+1}$ for 3D occupancy field $\hat{F}(x,y,z)$}. 

\fq{
\subsubsection{FOF with Cosine Series}
In FOF-X, we use the cosine series instead of the Fourier series as the basis for the subspace approximation.
The cosine series shares similar properties with the Fourier series but works with even periodic extension. Theoretically, even periodic extension can avoid discontinuities, preventing the overshoot known as the Gibbs phenomenon. \emph{Therefore, the cosine series version of FOF representation is used in FOF-X and throughout the following sections of this paper.} The only change is to modify Eq. (\ref{FS}) to:
\begin{equation}
  \label{CS}
  f_{ep}(z)= \frac{a_0}{2} + \sum_{n=1}^{\infty}a_n\cos (n\pi \frac{(z+1)}{2}).
\end{equation} 
Because $f(z)$ is defined on $[-1,1]$ but not $[0,1]$, a shift on $z$ is needed, which contributes to the term $\frac{(z+1)}{2}$ in Eq. (\ref{CS}). }
\yywy{We further analyze the representation efficiency of Fourier and cosine series under the same FOF formulation. As shown in the supplementary material, cosine series consistently achieve lower approximation error with fewer coefficients, motivating our design choice.}

\subsection{Image-to-Image Reconstruction Network}
\label{learn}
\yyw{In FOF-X, we \yyw{introduce} a novel reconstruction network with dual-sided normal map to mitigate lighting interference in real-world capture scenarios, coupled with integrated SMPL priors that enhance pose robustness under challenging articulations.}
\fq{\subsubsection{Reconstruction with Dual-sided Normal Maps} 
One important change in FOF-X is that we rely on dual-sided normal maps estimated from input RGB images rather than directly using the RGB images themselves to reconstruct human geometry. When testing with RGB input, we found a domain gap between the training images and real captured images, which significantly degraded the performance of our method. Although we try to mitigate this gap with a \yyw{physics-based} rendering engine (see Sec. \ref{datasets}), the reconstruction network is still sensitive to different textures and lighting conditions. To solve this, we design a new reconstruction framework with the dual-sided normal maps as the internal representation. 
Together with the FOF of the corresponding SMPL model, it is then fed to the network and used to reconstruct the human geometry. Since normal maps only keep the geometric information and are not affected by texture and lighting, the reconstruction network can focus on geometric information with better robustness and performance.
}

Thanks to the efficient dimensionality reduction with subspace approximation, the task of learning FOF from dual-sided normal maps can be essentially regarded as an image-to-image process. To this end, we exploit the image encoder 
\yyw{built}
on the HR-Net \cite{WangSCJDZLMTWLX19} framework for its outstanding fitting capability in various vision tasks. We use the weak perspective camera model in our implementation, in which FOF (and thus the reconstructed geometry) is naturally aligned with the input RGB image. Note that FOF can also be used with the normalized device coordinate space (NDC space) so that various camera models, including the perspective camera, are compatible with the framework.

\fq{
\subsubsection{SMPL Prior} Reconstructing a 3D object from a single image is a highly 
\yl{ill-posed}
problem. To address this, 3D priors, such as parametric body models, can be used to enhance the robustness of the model. Thanks to the 2D field nature of FOF, we propose a novel approach that leverages 3D geometric prior in 2D neural networks. We adopt the SMPL model as a prior and convert it to FOF representation with the proposed algorithm described in Sec. \ref{convertion1}. Then, we concatenate it with the RGB image as the input together. This can be written as:
\begin{equation}
    \bm{F_H} = G(\bm{N}_{dual},\bm{P}),
\end{equation}
where $G(\cdot)$ represents an image-to-image \yl{translation} neural network, which is HR-Net-48 in our implementation, $\bm{N}_{dual}$ \yl{denotes} the dual-sided normal maps, and $\bm{P}$ is the FOF with 16 terms of the corresponding SMPL model estimated from the input image.
The use of the prior makes the network more robust to different poses and 
\yl{significantly}
improves reconstruction quality for cases with high ambiguity. 
} 
\fq{
\subsubsection{Supervision}  To train our network, we sample points on the $xy$-plane and supervise the coefficients of the lines corresponding to these points with mean squared error (MSE) loss. To make the network more focused on human geometry, we only supervise the human foreground region of the image. The loss function $\mathcal{L}$ is formulated as: 
\begin{equation}
  \label{LOSS}
  \mathcal{L} = \frac{1}{n} \sum_{i=1}^n  \left\Vert \hat{\bm{C}}(x_i,y_i)-\bm{C}(x_i,y_i) \right\Vert_2^2,
\end{equation} 
where $(x_i,y_i) \in \mathcal{M}$ are sampled foreground pixels, $\mathcal{M}$ is the foreground region, $n$ is the number of foreground pixels, $\bm{C}(x_i,y_i)$ is the ground-truth FOF coefficient vector, and $\hat{\bm{C}}(x_i,y_i)$ is the FOF coefficient vector predicted by the network.
}

\begin{figure}[t]
\centering
\includegraphics[width=0.8\linewidth]{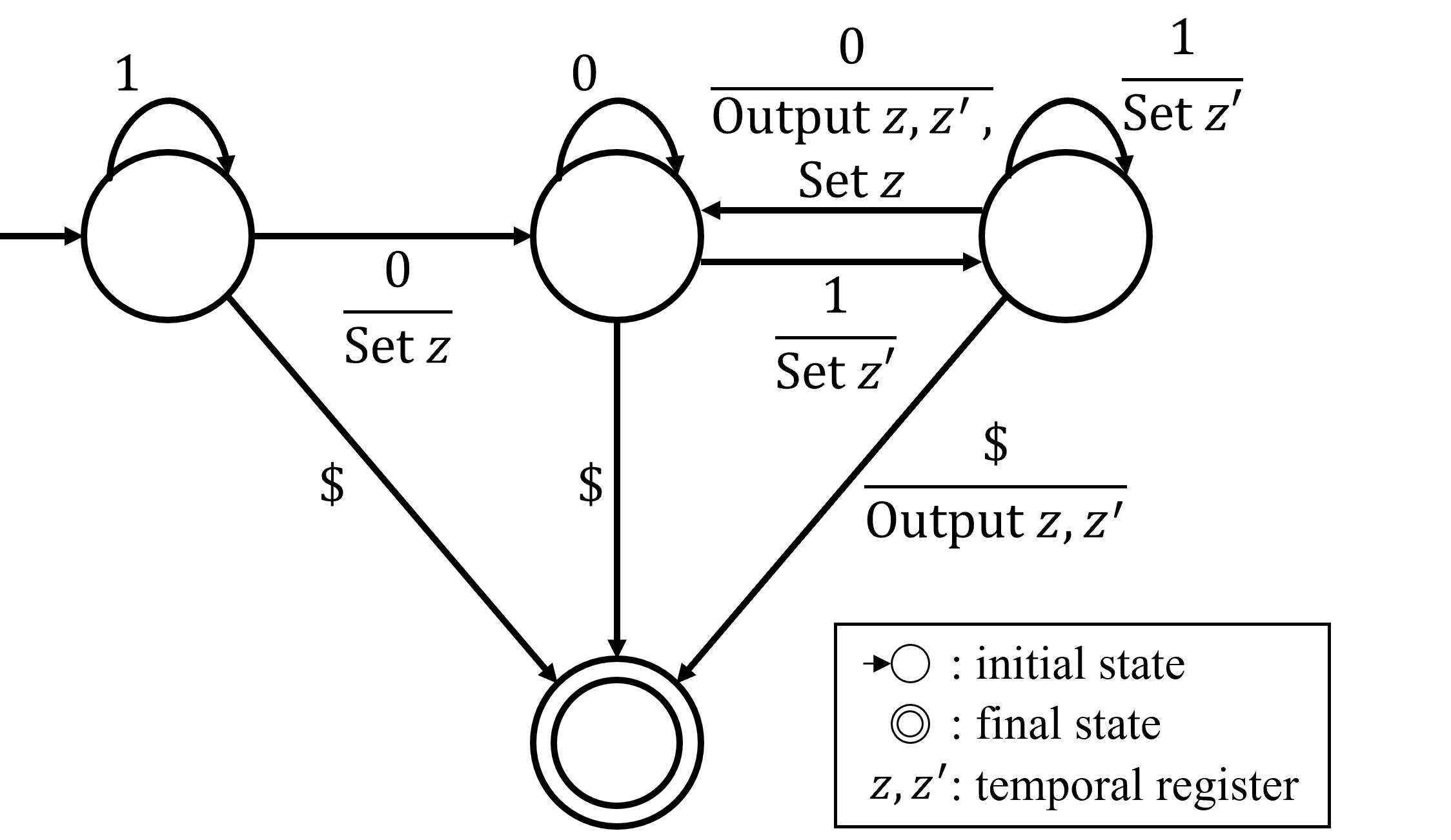}
\caption{\yywy{Automaton-based discontinuity matcher used to process discontinuities on each pixel. The results of with and without Automaton-based discontinuity matcher are shown in Fig. \ref{imp}.}} 
\label{automaton}
\end{figure}

\begin{figure}[t]
\centering
\includegraphics[width=.98\linewidth]{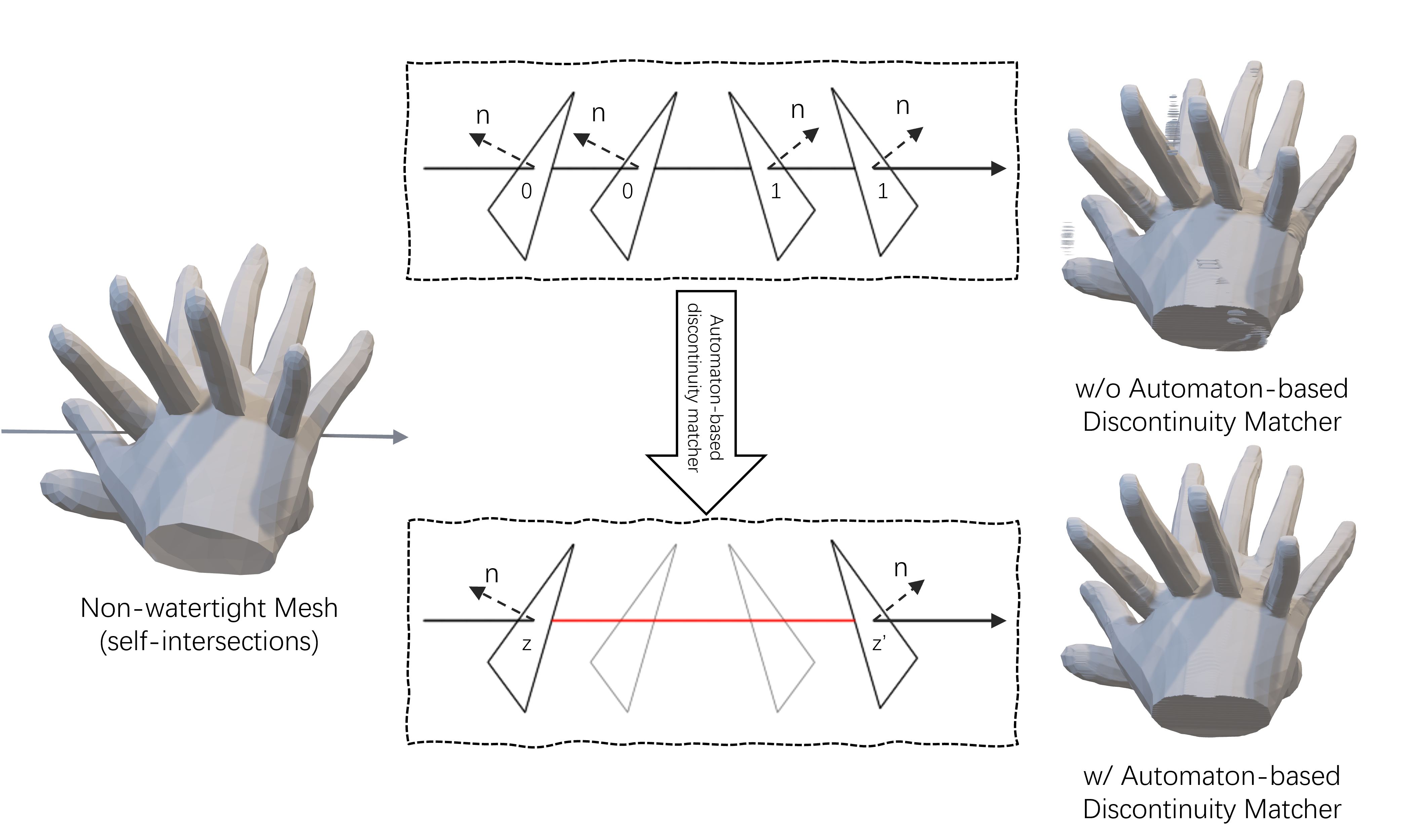}
\caption{\yywy{Example of Discontinuity Point Matching via Automaton.}}
\label{example}
\end{figure}

\subsection{Inter-Conversion between FOF and Meshes}
In FOF-X, the inter-conversion algorithms between FOF and \yl{mesh} representations are greatly improved.
\subsubsection{Mesh to FOF with Automaton-based Discontinuity Matcher}
\label{convertion1}
To prepare training data, 3D meshes are converted into FOF as training labels and input priors. According to Eq.~(\ref{CS}), this amounts to computing the first few terms of the cosine series for each 1D occupancy signal $f(z)$. Unlike generic signals requiring numerical integration, occupancy signals allow exact analytic solutions.

Formally, suppose the line associated with $f(z)$ intersects the mesh $k$ times, with $(z_i, z_i')$ denoting the entry and exit points. Then $\mathcal{Z}={(z_i,z_i')}{i=1}^k$ defines the inside intervals, and the cosine coefficients are
\begin{equation}
\label{SUM}
\begin{aligned}
a_n&= \int_{-1}^{1} f(z) \cos (t_n(z+1)) dz \\
   &= \frac{1}{t_n} \sum_{i=1}^k
   \sin \left( t_n \left(z_i'+1\right) \right) - 
   \sin\left( t_n \left(z_i+1\right)\right),
\end{aligned}
\end{equation}
where $t_n = \tfrac{n\pi}{2}$. Since $k$ is typically small (1–3 for human meshes), each coefficient can be computed in $O(1)$, and the whole mesh-to-FOF conversion runs in $O(HWN)$ with GPU parallelization, where $H,W$ are FOF resolution.

In practice, ground-truth meshes are not always watertight, leading to irregular discontinuity sequences (e.g., consecutive entries or exits). To solve this, we propose an automaton-based discontinuity matcher to enhance the robustness. Unlike what we do in the conference version~\cite{fof}, we collect those discontinuities and normal vectors for each line first and process them with an automaton before \yl{working out the integral}.
\yyw{For watertight meshes, both our automaton and the conference version produce identical results,  correctly converting meshes to FOF. For non-watertight meshes (e.g., 
surfaces with open boundaries or self-intersections, partly missing, or with non-manifold connectivity), the conference version method fails to compute FOF accurately, leading to floating artifacts (Fig. \ref{imp}). Our automaton-based discontinuity matcher robustly handles irregular patterns arising from non-watertight meshes (including holes, self-intersections, and duplicate faces), effectively preventing reconstruction artifacts caused by such topological imperfections. \yywy{We use the automaton described in Fig. \ref{automaton} to handle all the lights. For a more detailed approach please refer to section A of the supplementary material.} Fig. \ref{example} shows an example of our automaton processing a non-watertight mesh.}

\subsubsection{FOF to Mesh with Laplacian Coordinate Constraint}
\label{convertion2}
Extracting meshes is necessary for applications such as mesh-based rendering and animation. Recent methods based on implicit neural representations need to 
\yl{apply an MLP network}
on a 3D sampling grid. Instead, as described in Eq. (\ref{FOF}), the reconstruction of 3D occupancy field from FOF is simply a multiplication of two tensors. Then, the 3D mesh is extracted from the iso-surface of the 3D occupancy at the threshold of 0.5 with the Marching Cubes algorithm \cite{marching}. Both steps can be parallelized on GPUs for fast mesh generation.

\begin{figure}[!t]
\centering
\includegraphics[width=\linewidth]{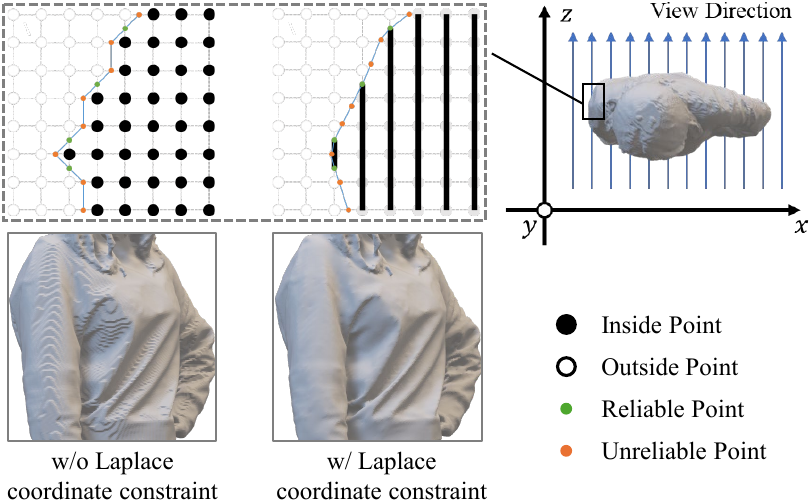}
\caption{Our FOF-to-mesh algorithm with Laplacian coordinate constraint. Compared with the original Marching Cubes, our conversion algorithm avoids the stair-step artifacts. }
\label{MC_pro}
\end{figure}

\fq{However, it is crucial to note that the occupancy field generated by FOF exhibits a view-dependent bias, resulting in observable stair-step artifacts on the reconstructed mesh. This bias marginally impacts the geometric accuracy and only detracts from the visual quality. As illustrated in Fig. \ref{MC_pro}, when applying Marching Cubes on FOF, only the points on the edges parallel to the $z$-axis remain precise. To address this, we propose a Laplacian coordinate constraint to eliminate the view bias.
\yl{We} adapt the Marching Cubes algorithm to incorporate an additional attribute for each point, which indicates the reliability of the point's coordinates. 
\yyw{The points produced on the edges parallel to the $z$-axis are reliable, and the points produced on the edges orthogonal to the $z$-axis are not.}
For those unreliable points, their coordinates are adjusted to confer greater mesh smoothness. This refinement is accomplished by minimizing the mean square of the Laplacian coordinates, expressed as:
\begin{equation}
\label{Lap}
\mathop{\arg\min}_{X_i, i\in U}  \left\Vert  (D-A)X  \right\Vert_2^2,
\end{equation}
where $D$ is a diagonal matrix with $D_{ii}=d_i$ on the diagonal, $d_i$ is the number of neighbors of the $i^{\textrm{th}}$ point, $X\in\mathbb{R}^{n\times 3}$ is \yl{the} coordinate matrix to be optimized, and $A$ is the adjacency matrix. $A_{ij} = 1$ if points $v_i$ and $v_j$ are adjacent, otherwise $A_{ij} = 0$. $U$ is the index set of unreliable points, which are to be optimized. 
\yyw{Unlike excessive Laplacian smoothing iterations that degrade geometric features into spherical approximations, our Laplacian coordinate constraint preserves topological fidelity and guarantees convergence to ground-truth mesh shapes.}

}

\begin{table*}[t]

  \caption{\yyw{Quantitative Comparison with existing methods. }}
  \label{comp}
  \centering
  \begin{tabular}{lll|ccccccccc}
    \toprule
    \multirow{3}{*}{} &&
    &\multicolumn{3}{c}{CAPE\cite{cape1}} 
    &\multicolumn{3}{c}{THuman2.1\cite{tao2021function4d}} 
    &\multicolumn{3}{c}{CustomHumans\cite{cus}}\\
    \cmidrule{4-12}
    & Year
    & Time$\downarrow$
    & Chamfer$\downarrow$   
    & P2S$\downarrow$
    & Normal$\downarrow$
    & Chamfer$\downarrow$ 
    & P2S$\downarrow$
    & Normal$\downarrow$
    & Chamfer$\downarrow$
    & P2S$\downarrow$
    & Normal$\downarrow$
    \\
    \midrule
    PIFu$^*$\cite{saito2019pifu}    & 2019 & 9.98  & 1.911 & 2.265 & 1.461 & 3.589 & 3.792 & 2.504 & 3.635 & 3.932 & 2.365\\
    PIFuHD \cite{saito2020pifuhd}   & 2020 & 15.90 & 3.279 & 3.393 & 1.903 & 3.963 & 3.986 & 2.620 & 4.184 & 4.115 & 2.432\\
    PaMIR$^*$ \cite{zheng2021pamir} & 2021 & 36.48 & 1.426 & 1.629 & 1.377 & 1.549 & 1.771 & 1.720 & 1.211 & 1.404 & 1.350\\
    ICON \cite{xiu2022icon}         & 2022 & 35.00 & 0.967 & 1.001 & 0.909 & 0.968 & 0.881 & 1.338 & 0.798 & 0.755 & 1.039\\
    ICON$_{fl}$ \cite{xiu2022icon}  & 2022 & 35.56 & 0.817 & 0.814 & 0.755 & 0.948 & 0.860 & 1.254 & 0.811 & 0.761 & 0.984\\
    D-IF \cite{yang2023dif} & 2023 & 61.15  & 0.758 & 0.714 & 0.752 & 1.210 & 1.082 & 1.332 & 1.043 & 0.955 & 1.098\\
    ECON$_{if}$ \cite{econ} & 2023 & 124.55 & 0.947 & 0.956 & 0.928 & 1.186 & 1.224 & 1.442 & 1.015 & 1.027 & 1.106\\
    ECON$_{ex}$\cite{econ}  & 2023 & 108.99 & 0.887 & 0.864 & 0.844 & 1.094 & 1.049 & 1.301 & 1.055 & 1.040 & 1.091\\
    GTA   \cite{gta}  & 2023 & 50.29 & 0.738 & 0.761 & 0.808 & 1.514 & 1.524 & 1.555 & 1.382 & 1.435 & 1.332\\
    SiFU \cite{sifu}  & 2024 & 62.38 & \bf{0.676} & 0.677 & 0.733 & 1.015 & 1.004 & 1.271 & 0.880 & 0.900 & 0.994\\
    SiTH \cite{sith}  & 2024 & 83.18 & 1.016 & 1.030 & 1.051 & 1.218 & 1.160 & 1.469 & 1.111 & 1.061 & 1.216\\
    PSHuman \cite{li2025pshuman}  & 2025 & 76.69 & 2.208 & 1.922 & 0.718 & 1.464 & 1.276 & \bf{0.675} & 2.066 & 1.815 & \underline{0.772}\\
    \midrule
    FOF-Base$^\dag$ \cite{fof} & 2022 & \underline{0.09} & 4.583 & 3.825 & 2.345 & 3.597 & 3.408 & 2.489 & 4.677 & 4.084 & 2.671\\
    FOF-SMPL$^\dag$ \cite{fof} & 2022 & - & 0.765 & 0.709 & 0.649 & 0.959 & 0.849 & 1.120 & 0.850 & 0.759 & 0.882\\
    FOF-X(HR32) (Ours) & - & \bf{0.02}
    & \underline{0.708} & \bf{0.654} & \bf{0.634}
    & \underline{0.858} & \underline{0.768} & 1.049
    & \underline{0.731} & \underline{0.663} & 0.783\\
    FOF-X(HR48) (Ours) & - & \bf{0.02}
    & 0.716 & \underline{0.672} & \underline{0.639}
    & \bf{0.835} & \bf{0.754} & \underline{1.033}
    & \bf{0.718} & \bf{0.658} & \bf{0.771}\\
    \bottomrule
  \end{tabular}
\end{table*}

\section{Experiments}
\subsection{Datasets}
\label{datasets}
We collect THuman2.1 \cite{tao2021function4d}, CustomHumans \cite{cus}, and CAPE \cite{cape1} for our experiments. To maintain reproducibility, all the datasets we use are publicly accessible. For fair comparisons, we follow ICON \cite{xiu2022icon} by training our method on THuman2.0, which is a subset of the first 526 items in THuman2.1, and using the CAPE dataset for evaluation. Additionally, we create two additional benchmarks on THuman2.1 and CustomHumans to evaluate our method and all the baselines.

To obtain RGB images, we render the meshes in Blender using 630 HDRi environment texture maps from Poly Haven\footnote{https://polyhaven.com/}. The HDRi maps serve both as panoramic background images and sources of ambient lighting. Each subject in the benchmarks is rendered to four images, uniformly distributed along the yaw axis with a fixed elevation of 0 degrees.  
For consistency with the CAPE dataset, all the meshes are normalized to a height of 1.8 meters. 
\yl{In practice,}
we find that \yl{the human body occupying a large percentage of the image and a clearer image} typically lead to better reconstruction results.

\begin{figure*}[t]
    \centering
    \includegraphics[width=0.98\linewidth]{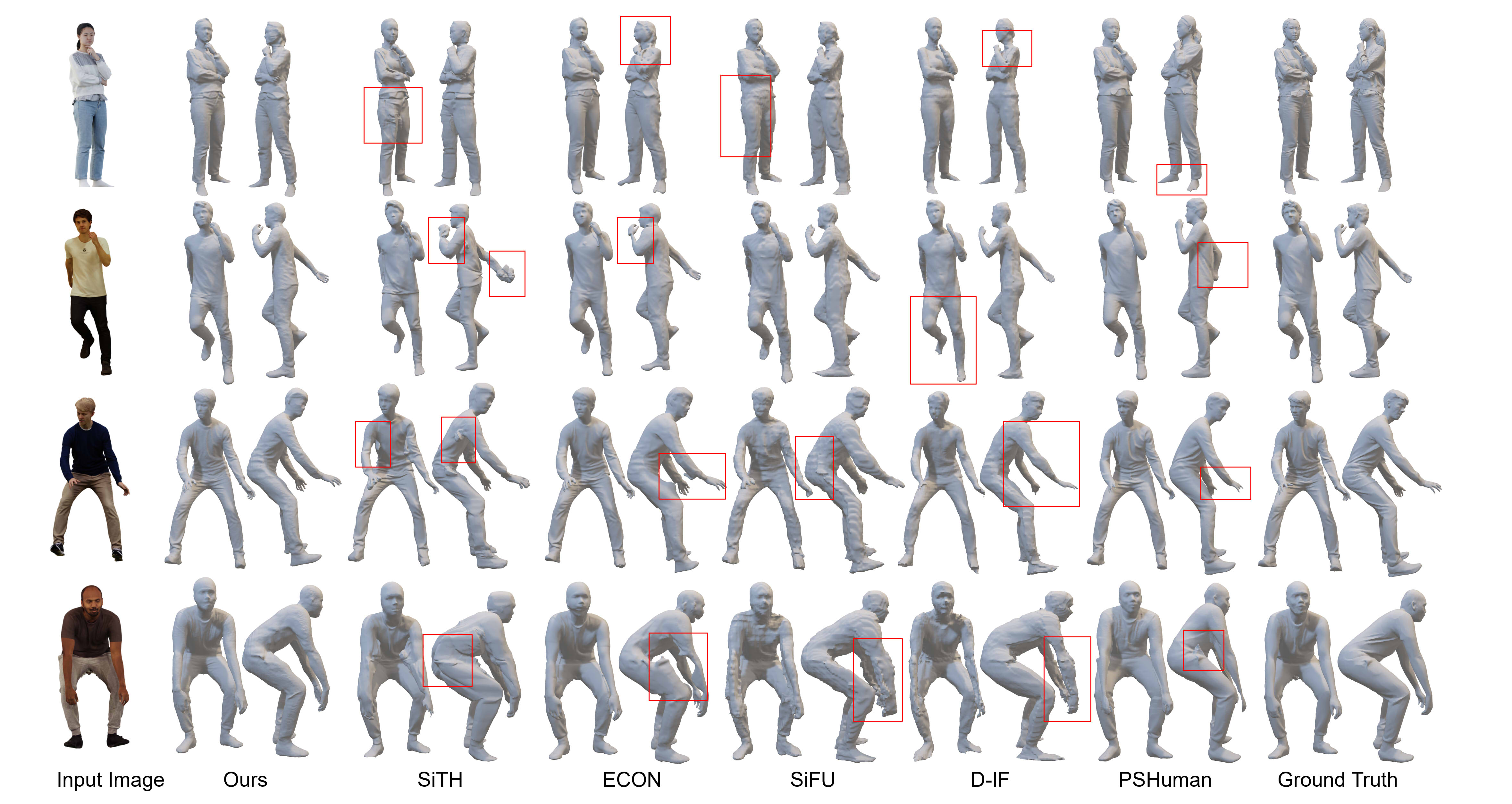}
    \caption{\yyw{Qualitative comparison against the state-of-the-art methods. }}
    \label{R_pose}
\end{figure*}

\subsection{Comparison with the State of the \yl{Arts}}

\label{comparison}
We compare our method with the state-of-\yl{the-}art approaches, including PIFu \cite{saito2019pifu}, PIFuHD \cite{saito2020pifuhd}, PaMIR\cite{zheng2021pamir}, ICON \cite{xiu2022icon}, ECON\cite{econ}, D-IF\cite{yang2023dif}, GTA\cite{gta}, SiFU\cite{sifu}, SiTH\cite{sith}, \yyw{and PSHuman\cite{li2025pshuman}} on CAPE\cite{cape1}, THuman2.1\cite{tao2021function4d}, and CustomHumans\cite{cus}.

\paragraph{Quantitative Results} 
The quantitative results are summarized in Table \ref{comp}. $^*$ denotes the re-implemented version in ICON\cite{xiu2022icon}. $^\dag$ denotes the re-implemented version with the cosine series and the improved inter-conversion algorithms. 
FOF-X(HR32) and FOF-X(HR48) are FOF-X model with HRNet-W32 and HRNet-W48 respectively.
\yyw{All methods listed, except PIFuHD\cite{saito2020pifuhd} and PSHuman\cite{li2025pshuman}, are trained on THuman2.0\cite{tao2021function4d}. It is worth noting that PSHuman's\cite{li2025pshuman} dataset includes THuman 2.1.} For PIFu\cite{saito2019pifu} and PaMIR\cite{zheng2021pamir}, similar to ICON\cite{xiu2022icon}, we use the re-implemented versions, as the official releases are trained with additional commercial datasets. 
Since some methods \yl{do not} provide the training sample lists, we use THuman2.1 instead of THuman2.0 to evaluate all methods.
We use $512\times 512\times 512$ resolution for the input of all these methods. We do not compare our method with \cite{alldieck2022phorhum} and \cite{He2021ARCHAC} because their codes are not released.  For simplicity, we also ignore some works that have already been compared with the state-of-the-art methods listed above.
We found that some methods are sensitive to the ratio of the human subject's size in the image. To ensure consistency, we use a ratio of 0.9 for all methods, corresponding to the 1.8m height normalization mentioned in the datasets section. 
Ground-truth SMPL meshes are used to better demonstrate the capability of the evaluated methods.

\yyw{As can be seen, our methods show the lowest reconstruction errors compared to other methods on CustomHumans datasets.}
As mentioned in \cite{sith}, CAPE provides the fitted SMPL+D meshes instead of the original scanned meshes as the ground-truth, leading to mismatches between the image details and the ground-truth meshes. This may cause a potential bias in evaluation. Despite this, our FOF-X can still achieve a comparable metric on CAPE. 
\yyw{In addition, on the THuman2.1 dataset, we are only slightly inferior to PSHuman\cite{li2025pshuman} in terms of normals. However, we note that PSHuman's\cite{li2025pshuman} training dataset incorporates THuman2.1 without providing clear train/test splits, potentially allowing some test data to be included in training.}
Notably, our method demonstrates robustness across all benchmarks, achieving better or comparable performance to other computationally heavy methods.


\paragraph{Qualitative Results}
We conduct a qualitative comparison in Fig. \ref{R_pose}. \yyw{As can be seen, SiTH\cite{sith} and PSHuman\cite{li2025pshuman} generally produces good results but sometimes introduces artifacts in the limbs and side shapes.}
ECON\cite{econ} suffers from self-occlusions which is the essential limitation of its optimization-based solution;
SiFU\cite{sifu} cannot recover detailed surfaces, although it can produce accurate rough shapes;
D-IF\cite{yang2023dif} has difficulties dealing with limb shapes due to the way it uses the SMPL prior.
In contrast, our method is able to reconstruct
plausible 3D human models under different body poses
and shapes. In terms of surface quality and geometric accuracy, our method outperforms other state-of-the-art approaches. Notably, our method is much more lightweight than others, allowing for real-time reconstruction.



\paragraph{Comparison of Running Times} Table \ref{comp} shows the comparison results in terms of running time.
\yyw{
All methods except PSHuman\cite{li2025pshuman} are tested on a single RTX-3090 GPU, where FOF-X represents the acceleration of FOF-X using TensorRT. PSHuman\cite{li2025pshuman} is tested on the A100 GPU due to memory limitations.} 
Our FOF-X takes 0.02 seconds per frame, demonstrating 500× acceleration over the fastest baseline (PIFu\cite{saito2019pifu}: 9.98s).
The original system presented in our conference publication, FOF\cite{fof}, is implemented in Python. Due to limitations in the original mesh-to-FOF algorithm, the SMPL prior is not supported in the FOF real-time system. Since FOF-SMPL requires the execution of three separate scripts, its running time is not reported. Our enhanced pipeline FOF-X adds additional components such as SMPL estimation, image segmentation, and rendering while maintaining real-time performance. Notably, all variants of our method (Python/TensorRT) consistently outperform existing \yyw{works in terms of efficiency} by two orders of magnitude.



\subsection{Ablation Study}
In this section, we conduct ablation experiments to validate the effectiveness of the settings and the proposed components in our FOF representation and FOF-X framework. Note that the ablation studies on the number of Fourier terms to use and the noise robustness of FOF are not using neural networks. 
These experiments are designed to explore the intrinsic properties of the FOF representation itself rather than the performance of the reconstruction algorithm.

\begin{figure}[t]
    \centering
    \includegraphics[width=0.95\linewidth]{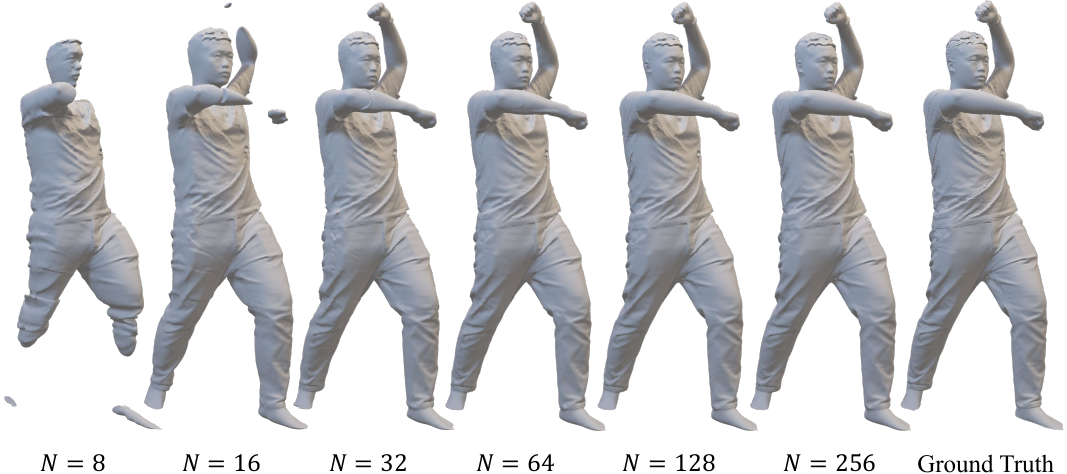}
    \caption{Results of different number $N$ of coefficients used in FOF-X.}
    \label{R_abla}
\end{figure}

\yywy{\subsubsection{Number of Terms to Use}
In the original FOF, we use $2N+1=31$ terms of the Fourier series to represent the occupancy field, which has been shown to be sufficient for maintaining geometric accuracy. However, we observe that the absence of high-frequency components can lead to visual artifacts. Therefore, we use $N=128$ terms of the cosine series in FOF-X. We illustrate the impact of varying $N$ on the accuracy of the approximation in Fig. \ref{R_abla}. The metrics for different values of $N$ are also presented in Table \ref{num of n}.
When $N \leq 16$, noticeable artifacts appear on the meshes. Noting the geometric details in the ear region, although using $N=32$ preserves most geometric information, we prefer $N=128$ in FOF-X to ensure higher visual quality and eliminate potential artifacts.}

\subsubsection{Sampling Scalability}
\begin{figure}[t]
    \centering
    \includegraphics[width=0.95\linewidth]{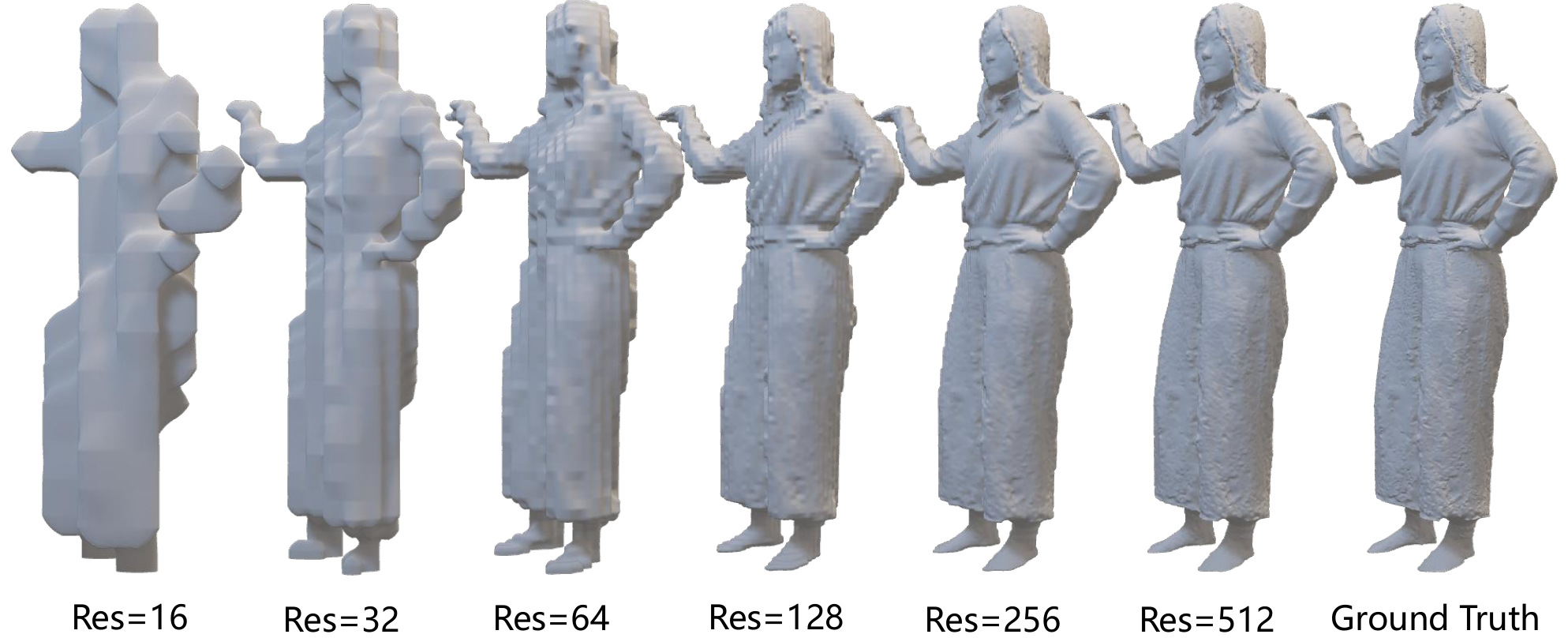}
    \caption{\yyw{Results of different resolution $Res$ used in FOF-X. We use $Res=256$ in our implementation.}}
    \label{R_res}
\end{figure}

The \yyw{frequency-domain} resolution of FOF can be adaptively inferred according to the time needs of the system, without the need for retraining. We explored the effect of resolution \yyw{(number of components)} on the results taking $N = 128$. Fig. \ref{R_res} illustrates the effect of changing resolution on the accuracy of the results. Metrics of different \yyw{$N$} values are also given in Table \ref{res}.


\begin{table}[t]
    \centering
        \caption{Geometric metrics for different numbers of FOF components.}
        \label{num of n}
    \begin{tabular}{ccccccc}
    
    \toprule
       $N$  & 8 & 16 & 32 & 64 & 128 & 256 \\
    \midrule
    P2S    & 1.342 & 0.466 & 0.148 & 0.054 & 0.027 & 0.024\\
    Chamfer  & 2.544 & 0.529 & 0.168 & 0.062 & 0.030 & 0.025\\
    \bottomrule
    \end{tabular}
\end{table}

\begin{table}[t]
    \centering
        \caption{\yyw{Geometric metrics for different resolution of FOF.}}
        \label{res}
    \begin{tabular}{ccccccc}
    
    \toprule
       $Res$  & 16 & 32 & 64 & 128 & 256 &512\\
    \midrule
    P2S    & 3.580 & 1.764 & 0.885 & 0.396 & 0.139 & 0.028\\
    Chamfer  & 3.655 & 1.746 & 0.889 & 0.403 & 0.146 & 0.031\\
    \bottomrule
    \end{tabular}
\end{table}

\begin{table}[t]
    \centering
        \caption{\yyw{Geometric Metrics for Different Mesh-to-FOF algorithms}}
        \label{mesh to fof}
    \resizebox{\linewidth}{!}{
    \begin{tabular}{cccccccc}
    \toprule
       & Chamfer$\downarrow$ &P2S$\downarrow$ &Normal$\downarrow$\\
    \midrule
    Algorithms Used in FOF     &0.447 &0.206 &0.264\\
    Algorithms Used in FOF-X   &\textbf{0.041} &\textbf{0.063} &\textbf{0.099}\\
    \bottomrule
    \end{tabular}}
\end{table}

\begin{table}[t]
    \centering
        \caption{\yyw{Visual Metrics for Different FOF-to-mesh algorithms}}
        \label{fof to mesh}
    \resizebox{\linewidth}{!}{
    \begin{tabular}{cccccccc}
    \toprule
       & SSIM$\uparrow$ &PSNR$\uparrow$ &LPIPS$\downarrow$\\
    \midrule
    w/o Laplacian
coordinate constraint     &0.959 &32.09 &0.018\\
    w/ Laplacian
coordinate constraint     &\textbf{0.965} &\textbf{32.35} &\textbf{0.015}\\
    \bottomrule
    \end{tabular}}
\end{table}

\subsubsection{Dual-sided Normal Maps}  
To mitigate the performance degradation caused by the domain gap between the training images and the real images, we take the RGB images as input and first map them to dual-sided normal maps for the subsequent pipeline. To verify the effectiveness of such a strategy, 
we compare FOF-X with our baseline method, which uses RGB images as input directly. As shown in Table \ref{comp}, with dual-sided normal maps as an internal representation, FOF-X-32 is more robust than the original FOF (FOF-SMPL$^\dag$) on test sets. 
Note that FOF-X is trained on ground-truth normal maps and tested on maps generated by an off-the-shelf model. Fig. \ref{teaser} shows the visual results.

\subsubsection{Improved Inter-Conversion Algorithm} Fig. \ref{imp} shows the effectiveness and robustness of our improved inter-conversion algorithm. As can be seen, our parallelized mesh-to-FOF algorithm demonstrates strong robustness. \yyw{The quantitative results for two mesh-to-FOF algorithms are also given in Table \ref{mesh to fof}.}
With the automaton-based discontinuity matcher, it effectively avoids the floating masses caused by duplicate or missing discontinuities. This algorithm is integrated into the real-time pipeline for its excellent parallelizability. 
\yyw{Table \ref{fof to mesh} shows the quantitative results of the FOF-to-mesh methods. To better represent the visual effect of the mesh, we render the mesh as a normal diagram and introduce visual metrics\footnote{SSIM (structural similarity), PSNR (peak signal-to-noise ratio) and LPIPS (learned perceptual image patch similarity) measure structural preservation, pixel-level accuracy and human perceptual consistency, respectively.} to evaluate results.} Our improved FOF-to-mesh algorithm avoids the stair-step artifacts caused by view bias effectively. Although the Laplacian coordinate constraint solution is a little time-consuming, we can use Laplacian smoothing in the real-time pipeline as an alternative and apply our full FOF-to-mesh algorithm during the mesh export stage. 

\section{Conclusion and Discussion}
In this paper, We presented the Fourier Occupancy Field (FOF), an efficient 3D representation for monocular real-time human reconstruction, and extended it to the FOF-X framework for higher-quality results.
FOF encodes the occupancy field into a compact multi-channel 2D form, enabling simple image-to-image estimation and seamless conversion with meshes.
Building on this, we proposed the FOF-X framework with enhanced conversion and reconstruction algorithms. Specifically, an automaton-based discontinuity matcher enables robust mesh-to-FOF conversion and incorporation of SMPL priors into real-time pipelines, while Laplacian coordinate constraints eliminate stair-step artifacts in FOF-to-mesh conversion. For reconstruction, we employ dual-sided normal maps as input, effectively reducing domain gaps between training and real images. Experiments on scanned and in-the-wild datasets confirm that FOF and FOF-X achieve high-quality and detailed human reconstruction, surpassing prior approaches.

The FOF representation is based on the Fourier series expansion of the square-wave-like function. When applying FOF to thin objects, the spectrum of the function will contain many high-frequency components. 
In such cases, we need more terms in the Fourier series to approximate the function. The number of terms we need is roughly inversely proportional to the thickness of the object. 
In our experiments, we found that FOF fails to reconstruct objects that are too thin, like single layer mesh. 
Although wavelet transformation and short-time Fourier transformation seem to be a promising option, the implementation and computational efficiency still need more exploration.
\yyw{In addition, the FOF representation shows significant potential for broader applications. Its mathematical formulation and efficient parameterization could be extended to general object representation and even scene-level modeling through appropriate architectural modifications. These promising directions will be explored in future work to enhance the framework's capabilities and applicability.}
We leave this issue for future work to further improve our framework.

Although a previous demo paper of FOF \cite{demo} has demonstrated a system with a perspective camera setting, it encountered challenges in handling changes in the camera parameters.
Integrating the camera setting of a true perspective camera with our method for more accurate mesh recovery remains a topic for future research.
Moreover, we have noticed that the combination of various 3D representations and generative models has produced a bunch of amazing works. Given the good nature of FOF, combining FOF with large models to sense the general 3D world is also a promising direction.
\bibliographystyle{IEEEtran}
\bibliography{IEEEabrv, citations}

%
\vspace{-1.15cm}
\begin{IEEEbiography}[{\includegraphics[width=1in,height=1.25in,clip,keepaspectratio]{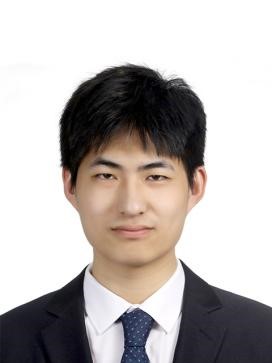}}]
{Qiao Feng} (Student Member, IEEE) received his B.Eng and M.Eng degrees in computer science from Tianjin University, Tianjin, China, in 2021 and 2024, respectively. His research interests include 3D Vision, with a current focus on 3D humans.
\end{IEEEbiography}

\vspace{-1.25cm}
\begin{IEEEbiography}[{\includegraphics[width=1in,height=1.25in,clip,keepaspectratio]{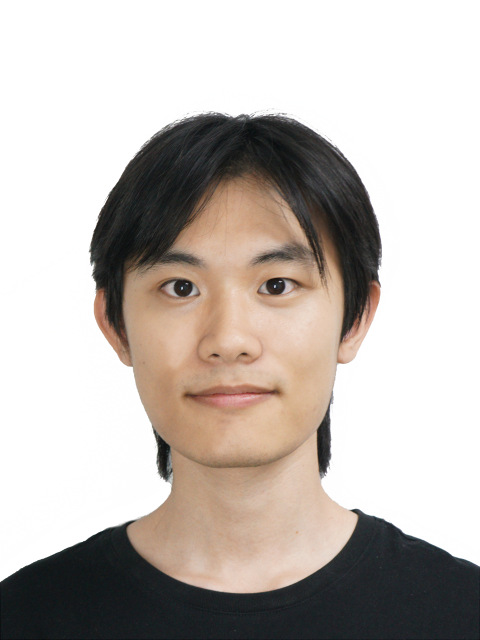}}]
{Yuanwang Yang} (Student Member, IEEE) received the B.E. degree from Hebei University of Technology, Tianjin, China, in 2023. He is currently pursuing the M.E. degree in computer science in Tianjin University. His research interests include computer vision, 3D and computer graphics.
\end{IEEEbiography}

\vspace{-1.25cm}
\begin{IEEEbiography}[{\includegraphics[width=1in,height=1.25in,clip,keepaspectratio]{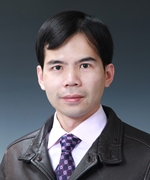}}]
{Yebin Liu} (Member, IEEE)
received the B.E. degree from the Beijing University of Posts and Telecommunications, China, in 2002 and the Ph.D. degree from the Department of Automation, Tsinghua University, Beijing, China, in 2009. He is currently a full Professor with Tsinghua University. He was a research fellow in the Computer Graphics Group of the Max Planck Institute for Informatik, Germany, in 2010. His research areas include computer vision, computer graphics, and computational photography.
\end{IEEEbiography}

\vspace{-1.25cm}
\begin{IEEEbiography}[{\includegraphics[width=1in,height=1.25in,clip,keepaspectratio]{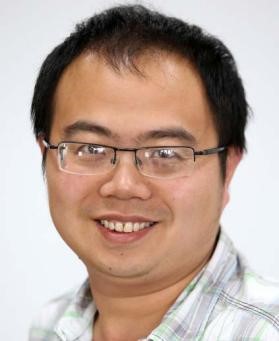}}]
{Yu-Kun Lai}
(Senior Member, IEEE) received his bachelor's and Ph.D. degrees in computer science from Tsinghua University in 2003 and 2008, respectively. He is currently a professor in the School of Computer Science \& Informatics, Cardiff University. His research interests include computer graphics, geometry processing, image processing and computer vision. He is on the editorial boards of IEEE Transactions on Visualization and Computer Graphics and The Visual Computer.
\end{IEEEbiography}

\vspace{-1.25cm}
\begin{IEEEbiography}[{\includegraphics[width=1in,height=1.25in,clip,keepaspectratio]{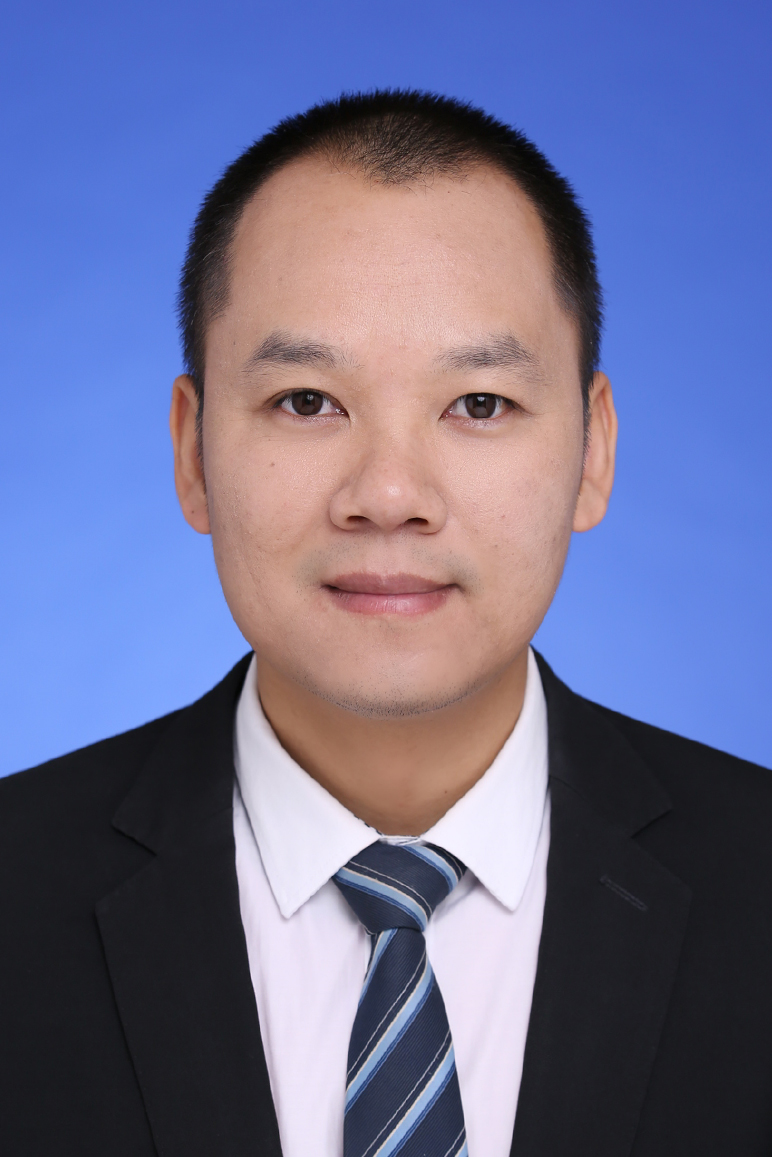}}]
{Jingyu Yang} (Senior Member, IEEE) received the B.E. degree from the Beijing University of Posts and Telecommunications, Beijing, China, in 2003, and the Ph.D. degree (Hons.) from Tsinghua University, Beijing, in 2009. He has been a Faculty Member with Tianjin University, Tianjin, China, since 2009, where he is currently a Professor with the School of Electrical and Information Engineering. He served as the Special Session Chair in VCIP 2016 and the Area Chair in ICIP 2017. His research focuses on image processing, 3-D imaging, and computer vision.
\end{IEEEbiography}

\vspace{-1cm}
\begin{IEEEbiography}[{\includegraphics[width=1in,height=1.25in,clip,keepaspectratio]{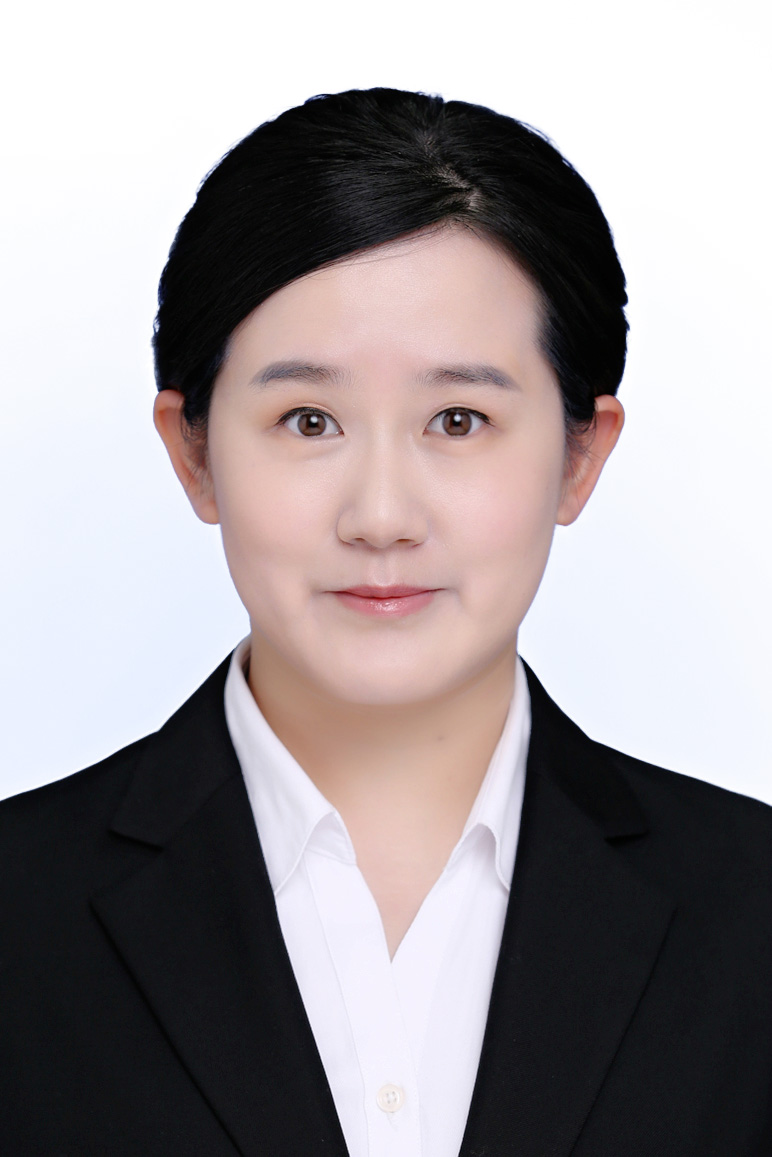}}]
{Kun Li}
(Senior Member, IEEE) received the B.E. degree from Beijing University of Posts and Telecommunications, Beijing, China, in 2006, and the master and Ph.D. degrees from Tsinghua University, Beijing, in 2011. She is currently a Professor with the College of Intelligence and Computing, Tianjin University, Tianjin, China. She was the recipient of the CSIG Shi Qingyun Award for Women Scientists. Her research interests include 3D reconstruction and AIGC.
\end{IEEEbiography}
\vfill

\end{document}